\documentclass[10pt,twocolumn,letterpaper]{article}

\usepackage{cvpr}
\usepackage{times}
\usepackage{epsfig}
\usepackage{graphicx}
\usepackage{amsmath}
\usepackage{amssymb}

\usepackage{mathtools}
\DeclarePairedDelimiter\ceil{\lceil}{\rceil}
\DeclarePairedDelimiter\floor{\lfloor}{\rfloor}

\usepackage[dvipsnames]{xcolor}

\usepackage{booktabs}
\usepackage{xcolor}
\usepackage{subcaption}
\usepackage{caption}
\usepackage[pagebackref=true,breaklinks=true,letterpaper=true,colorlinks=true, citecolor=RoyalBlue, bookmarks=false]{hyperref}

\cvprfinalcopy 

\def\cvprPaperID{8381} 

\begin{document}

\title{CenterMask : Real-Time Anchor-Free Instance Segmentation}

\author{Youngwan Lee and Jongyoul Park\thanks{Corresponding author.}\\
Electronics and Telecommunications Research Institute (ETRI), South Korea\\
{\tt\small \{yw.lee, jongyoul\}@etri.re.kr}
}

\maketitle

\begin{abstract}
   We propose a simple yet efficient anchor-free instance segmentation, called CenterMask, that adds a novel spatial attention-guided mask (SAG-Mask) branch to anchor-free one stage object detector (FCOS~\cite{Tian_2019_ICCV}) in the same vein with Mask R-CNN~\cite{he2017mask}.
   Plugged into the FCOS object detector, the SAG-Mask branch predicts a segmentation mask on each detected box with the spatial attention map that helps to focus on informative pixels and suppress noise.
   We also present an improved backbone networks, VoVNetV2, with two effective strategies: (1) residual connection for alleviating the optimization problem of larger VoVNet~\cite{lee2019energy} and (2) effective Squeeze-Excitation (eSE) dealing with the channel information loss problem of original SE. 
   With SAG-Mask and VoVNetV2, we deign CenterMask and CenterMask-Lite that are targeted each to large and small models, respectively.
   Using the same ResNet-101-FPN backbone, CenterMask achieves 38.3\%, surpassing all previous state-of-the-art methods while at a much faster speed.
   CenterMask-Lite also outperforms the state-of-the-art by large margins at over 35fps on Titan Xp.
   We hope that CenterMask and VoVNetV2 can serve as a solid baseline of real-time instance segmentation and backbone network for various vision tasks, respectively. 
   The Code is available at \url{https://github.com/youngwanLEE/CenterMask}.
\end{abstract}


\section{Introduction}
Recently, instance segmentation has made great progress beyond object detection.
The most representative method, Mask R-CNN~\cite{he2017mask}, extended on object detection (e.g., Faster R-CNN~\cite{ren2015faster}), has dominated COCO~\cite{lin2014microsoft} benchmarks since instance segmentation can be easily solved by detecting objects and then predicting pixels on each box.
However, even if there have been many works~\cite{huang2019mask,cai2018cascade,chen2019hybrid,li2019scale,liu2018path} for improving the Mask R-CNN~\cite{he2017mask}, few works exist for considering the speed of the instance segmentation. 
Although YOLACT~\cite{Bolya_2019_ICCV} is the first real-time one-stage instance segmentation due to its parallel structure and extremely lightweight assembly process, the accuracy gap from Mask R-CNN~\cite{he2017mask} is still significant. Thus, we aim to bridge the gap by improving both accuracy and speed.

\begin{figure}[t]
\centering
   \scalebox{0.45}{
      \includegraphics[width=\textwidth]{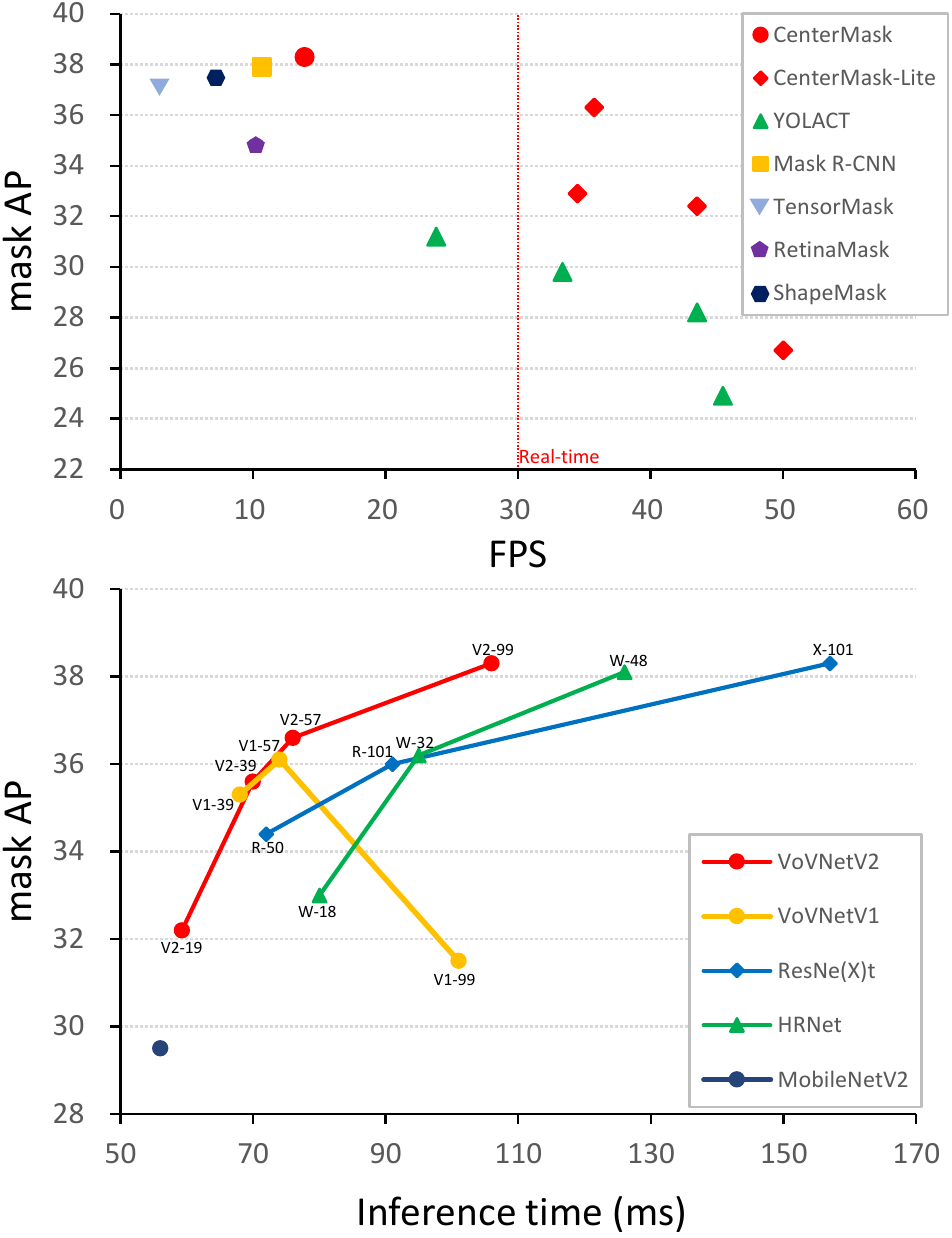} 
   }
\caption{\textbf{ Accuracy-speed Tradeoff.} across various instance segmentation models~(top) and backbone networks~(bottom) on COCO.
The inference speed of CenterMask \& CenterMask-Lite is reported on the same GPU (V100/Xp) with their counterparts.
Note that all backbone networks in the bottom are compared under the proposed CenterMask.
Please refer to section~\ref{sec:SOTA}, Table~\ref{tab:backbone} and Table~\ref{tab:SOTA} for details.}
\label{fig:FPS}
\vspace{-0.5cm}
\end{figure}

\begin{figure*}[t]
\centering
   \includegraphics[width=\textwidth]{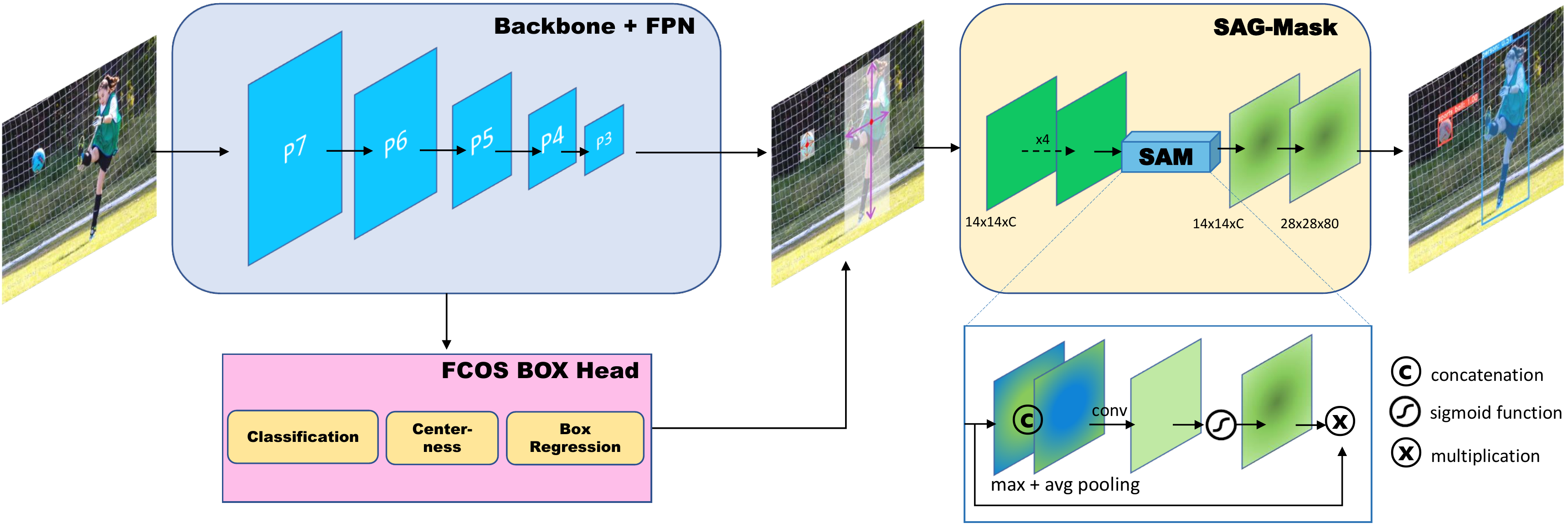} 
\caption{\textbf{Architecture of CenterMask.} where P3~(stride of $2^3$) to P7~(stride of $2^7$) denote the feature map in feature pyramid of backbone network. Using the features from the backbone, FCOS~\cite{Tian_2019_ICCV} predicts bounding boxes. Spatial Attention-Guided Mask~(SAG-Mask) predicts segmentation mask inside of the each detected box with Spaital Attention Module~(SAM) helping to focus on the informative pixels but also suppress the noise.}
\label{fig:architecture}
\vspace{-0.5cm}
\end{figure*}

While Mask R-CNN~\cite{he2017mask} is based on a two-stage object detector~(e.g., Faster R-CNN) that first generates box proposals and then predicts box location and classification, YOLACT~\cite{Bolya_2019_ICCV} is built on one-stage detector~(RetinaNet~\cite{lin2018focal}) that directly predicts boxes without proposal step.
However, these object detectors rely heavily on pre-define anchors, which are sensitive to hyper-parameters (e.g., input size, aspect ratio, scales, etc.) and different datasets. 
Besides, since they densely place anchor boxes for higher recall rate, the excessively many anchor boxes cause the imbalance of positive/negative samples and higher computation/memory cost.
To cope with these drawbacks of anchor boxes, recently, many works~\cite{law2018cornernet,Duan_2019_ICCV,zhou2019objects,zhou2019bottom,Tian_2019_ICCV,zhou2019objects} tend to escape from the anchor boxes toward \textit{anchor-free} by using corner/center points, which leads to more computation-efficient and better performance compared to anchor box based detectors.

Therefore, we design a simple yet efficient anchor-free one stage instance segmentation called \textit{CenterMask} that adds a novel spatial attention-guided mask branch to the more efficient one-stage anchor-free object detector~(FCOS~\cite{Tian_2019_ICCV}) in the same way with Mask R-CNN~\cite{he2017mask}. 
Figure ~\ref{fig:architecture} shows the overview of our CenterMask.
Plugged into the FCOS~\cite{Tian_2019_ICCV} object detector, our spatial attention-guided mask~(SAG-Mask) branch takes the predicted boxes from the FCOS~\cite{Tian_2019_ICCV} detector to predict segmentation masks on each Region of Interest~(RoI).
The spatial attention module~(SAM) in the SAG-Mask helps the mask branch to focus on meaningful pixels and suppressing uninformative ones.  

When extracting features on each RoI for mask prediction, each RoI pooling should be assigned considering the RoI scales. 
Mask R-CNN~\cite{he2017mask} proposes a new assignment function, called RoIAlign, that does not consider the input scale.
Thus, we design a scale-adaptive RoI assignment function that considers the input scale and is a more suitable one-stage object detector.
We also propose a more effective backbone network \textit{VoVNetV2} based on VoVNet~\cite{lee2019energy} that shows better performance and faster speed than ResNet~\cite{he2016deep} and DenseNet~\cite{huang2017densely} due to its One-shot Aggregation~(OSA).
In Figure~\ref{fig:FPS} (bottom), We found that stacking the OSA modules in VoVNet makes the performance degradation~(e.g., VoVNetV1-99). We see this phenomenon as the motivation of ResNet~\cite{he2016deep} because the backpropagation of gradient is disturbed.
Thus, we add the residual connection~\cite{he2016deep} into each OSA module to ease optimization, which makes the VoVNet deeper and in turn, boosts the performance.

In the Squeeze-Excitation (SE)~\cite{hu2018squeeze} channel attention module, it was found that the fully connected layers reduce the channel size, thereby reducing computational burden and unexpectedly causing channel information loss.
Thus, we re-design the SE module as \textit{effective} SE~(eSE) replacing the two FC layers with one FC layer maintaining channel dimension, which prevents the information loss and in turn, improves the performance.
With residual connection and eSE modules, We propose VoVNetV2 on various scales; from lightweight VoVNetV2-19, base VoVNetV2-39/57 and large model VoVNetV2-99 that are correspond with MobileNet-V2~\cite{howard2017mobilenets}, ResNet-50/101~\cite{he2016deep} \& HRNet-W18/32~\cite{sun2019high}, and ResNeXt-32x8d~\cite{xie2017aggregated}.

With SAG-Mask and VoVNetV2, we design CenterMask and CenterMask-Lite that are targeted each to large and small models, respectively.
The Extensive experiments demonstrate the effectiveness of CenterMask \& CenterMask-Lite and VoVNetV2.
Using the same ResNet-101 backbone~\cite{he2016deep}, CenterMask outperforms all previous state-of-the-art single models on the COCO~\cite{lin2014microsoft} instance and detection tasks while at a much faster speed.
CenterMask-Lite with VoVNetV2-39 bakcbone also achieves 33.4\% mask AP / 38.0\% box AP, outperforming the state-of-the-art real-time instance segmentation YOLACT~\cite{Bolya_2019_ICCV} by 2.6 / 7.0 AP gain, respectively, at over 35fps on Titan Xp.



\section{CenterMask}
In this section, first, we review the anchor-free object detector, FCOS~\cite{Tian_2019_ICCV}, which is a fundamental object detection part of our CenterMask. 
Next, we demonstrate the architecture of the CenterMask and describe how the proposed spatial attention-guided mask branch (SAG-Mask) is designed to plug into the FCOS~\cite{Tian_2019_ICCV} detector. 
Finally, a more effective backbone network, VoVNetV2, is proposed to boost the performance of CenterMask in terms of accuracy and speed.

\subsection{FCOS}
FCOS~\cite{Tian_2019_ICCV} is an anchor-free and proposal-free object detection in a per-pixel prediction manner as like FCN~\cite{long2015fully}.
Almost state-of-the-art object detectors such as Faster R-CNN~\cite{ren2015faster}, YOLO~\cite{redmon2016you}, and RetinaNet~\cite{lin2018focal} use the concept of the pre-defined anchor box which needs elaborate parameter tunning and complex calculation associated with box IoU in training.
Without the anchor-box, the FCOS~\cite{Tian_2019_ICCV} directly predicts a 4D vector plus a class label at each spatial location on a level of feature maps.
As shown in Figure ~\ref{fig:architecture}, the 4D vector embeds the relative offsets from the four sides of a bounding box to the location~(e.g., left, right, top and bottom).
In addition, FCOS~\cite{Tian_2019_ICCV} introduces the centerness branch to predict the deviation of a pixel to the center of its corresponding bounding box, which improves the detection performance.
Avoiding complex computation of anchor-boxes, FCOS~\cite{Tian_2019_ICCV} reduces memory/computation cost but also outperforms the anchor box based object detectors.
Because of the efficiency and good performance of the FCOS~\cite{Tian_2019_ICCV}, we design the proposed CenterMask built upon the FCOS~\cite{Tian_2019_ICCV} object detector.
\subsection{Architecture}
Figure \ref{fig:architecture} shows overall architecture of the  CenterMask. 
CenterMask consists of three-part:(1) backbone for feature extraction, (2) FCOS~\cite{Tian_2019_ICCV} detection head, and (3) mask head.
The procedure of masking objects is composed of detecting objects from the FCOS~\cite{Tian_2019_ICCV} box head and then predicting segmentation masks inside the cropped regions in a per-pixel manner.

\subsection{Adaptive RoI Assignment Function}\label{sec:3.2}
After object proposals are predicted in the FCOS~\cite{Tian_2019_ICCV} box head, CenterMask predicts segmentation masks using the predicted box regions in the same vein as Mask R-CNN.
As the RoIs are predicted from different levels of feature maps in Feature Pyramid Network (FPN~\cite{lin2017feature}), RoI Align~\cite{he2017mask} that extracts features should be assigned at different scales of feature maps with respect to RoI scales.
Specifically, an RoI with a large scale has to be assigned to a higher feature level and vice versa.
Mask R-CNN~\cite{he2017mask} based two-stage detector uses Equation \ref{eq:1}~ in FPN~\cite{lin2017feature} to determine which feature map (P\textsubscript{\textit{k}}) to be assigned.

\begin{equation} \label{eq:1}
k = \floor{k_0 + \log_2 \sqrt{wh}/224},
\end{equation}

\noindent
where \textit{k}\textsubscript{0} is 4 and \textit{w, h} are the width and height of the each RoI.
However, Equation~\ref{eq:1} is not suitable for CenterMask based one-stage detector because of two reasons.
First, Equation \ref{eq:1} is tuned to two-stage detectors (e.g.,FPN~\cite{lin2017feature}) that use different feature levels compared to one-stage detectors (e.g, FCOS~\cite{Tian_2019_ICCV}, RetinaNet~\cite{lin2018focal}).
Specifically, two-stage detectors use feature levels of P2 (stride of 2$^2$) to P5 (2$^5$) while one-stage detectors use from P3 (2$^3$) to P7 (2$^7$) that is larger receptive fields with lower-resolution.
Besides, the canonical ImageNet pretraining size 224 in Equation \ref{eq:1} is hard-coded and not adaptive to feature scale variation.
For example, when the input dimension is 1024$\times$1024 and the area of an RoI is 224$^2$, the RoI is assigned to relative higher feature P4 despite its small size of the area with respect to input dimension, which results in reducing small object AP.
Therefore, we define Equation \ref{eq:2} as a new RoI assignment function suited for CenterMask based one-stage detectors.

\begin{equation} \label{eq:2}
k = \ceil{k\textsubscript{max} - \log_2 \mathnormal{A}_{input}/\mathnormal{A}_{RoI}},
\end{equation}

\noindent
where \textit{k}\textsubscript{max} is the last level (e.g., 7) of feature map in backbone and $\mathnormal{A}_{input}$, $\mathnormal{A}_{RoI}$ are area of input image and the RoI, respectively. 
Without the canonical size 224 in Equation~\ref{eq:1}, Equation~\ref{eq:2} adaptively assign RoI pooling scale by the ratio of input/RoI area. 
If \textit{k} is lower than minimum level (e.g., P3), \textit{k} is clamped to the minimum level. 
Specifically,  if the area of an RoI is bigger than half of the input area, the RoI is assigned to the highest feature level(e.g., P\textsubscript{7}). 
Inversely, while Equation~\ref{eq:1} assigns P\textsubscript{4} to the RoI with 224$^2$, Equation~\ref{eq:2} determine \textit{k}\textsubscript{max} - 5 level which maybe minimum feature level for area of the RoI that is about $\times20$ smaller than input size. 
We can find that the proposed RoI assignment method improves the small object AP than Equation~\ref{eq:1} because of its adaptive and scale-aware assignment strategy in Table~\ref{tab:level}.
From an ablation study, we set \textit{k}\textsubscript{max} to P5 and \textit{k}\textsubscript{min} to P3.

\begin{figure*}[t]
\centering
\scalebox{0.9}{
   \includegraphics[width=\textwidth]{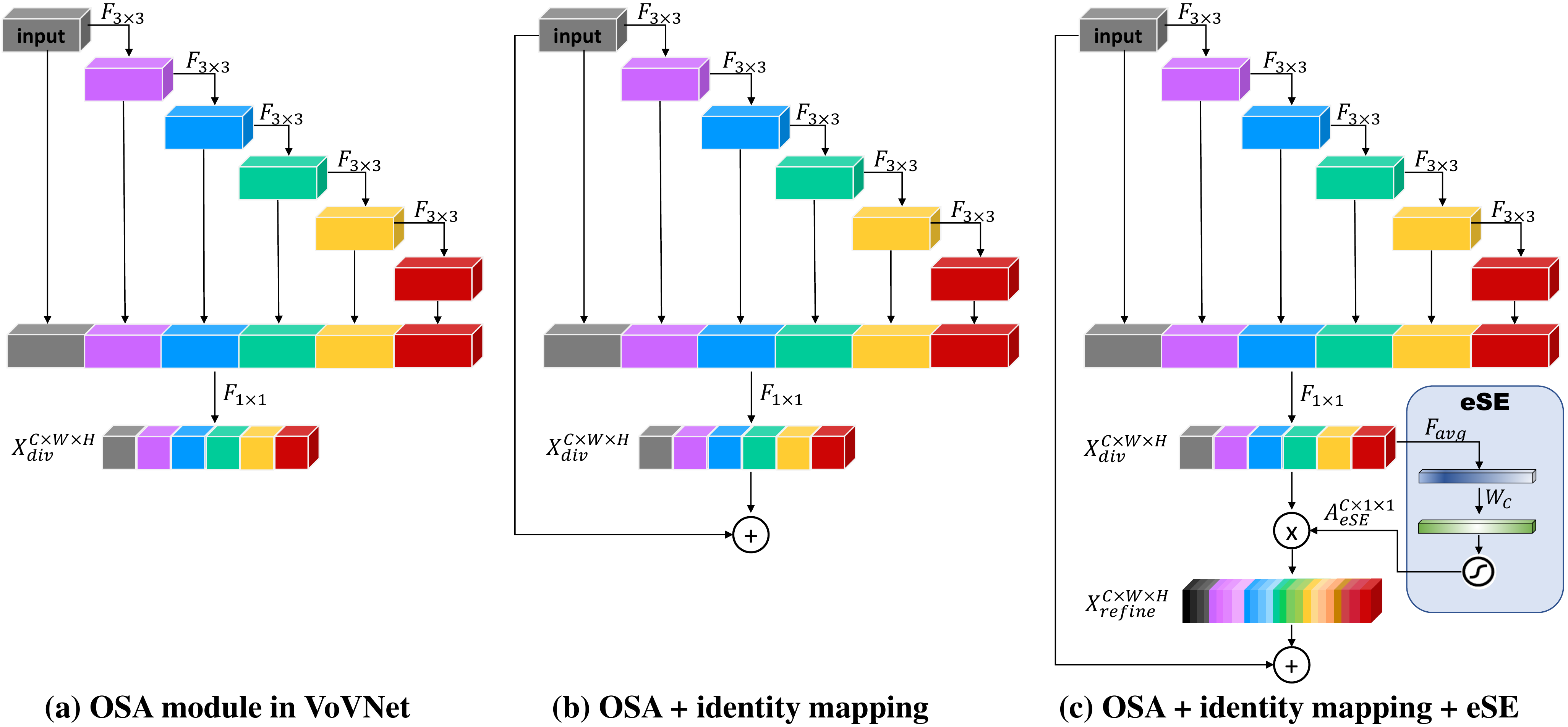} 
}
\caption{Comparison of OSA modules. $F_{1\times1}, F_{3\times3}$ denote $1\times1, 3\times3$ \texttt{conv} layer respectively, $F_{avg}$ is global average pooling, $W_C$ is fully-connected layer, $A_{eSE}$ is channel attention map, $\otimes$ indicates element-wise multiplication and $\oplus$ denotes element-wise addition.}
\label{fig:osa}
\vspace{-0.5cm}
\end{figure*}

\subsection{Spatial Attention-Guided Mask}
Recently, attention methods~\cite{hu2018squeeze,woo2018cbam,Zhu_2019_ICCV,Qin_2019_ICCV} have been widely applied to object detections because it helps to focus on important features, but also suppress unnecessary ones.
In particular, channel attention~\cite{hu2018squeeze,hu2018gather} emphasizes `what' to focus across channels of feature maps while spaital attention~\cite{woo2018cbam,chen2017sca} focuses `where' is an informative regions.
Inspired by the spatial attention mechanism, we adopt a spatial attention module to guide the mask head for spotlighting meaningful pixels and repressing uninformative ones.

Thus, we design a spatial attention-guided mask (SAG-Mask), as shown in Figure \ref{fig:architecture}.
Once features inside the predicted RoIs are extracted by RoI Align~\cite{he2017mask} with 14$\times$14 resolution, those features are fed into four \texttt{conv} layers and spatial attention module (SAM) sequentially.
To exploit the spatial attention map $\mathrm{A}_{sag}(\mathrm{X}_i) \in \mathbb{R}^{1 \times W \times H}$ as a feature descriptor given input feature map $\mathrm{X}_i \in \mathbb{R}^{C\times W \times H}$, the SAM first generates pooled features $\mathrm{P}_{avg}$, $\mathrm{P}_{max} \in \mathbb{R}^{1 \times W \times H} $  by both average and max pooling operations respectively along the channel axis and aggregates them via concatenation.
Then it is followed by a $3\times3$ \texttt{conv} layer and normalized by the sigmoid function. The computation process is summarized as follow:

\begin{equation} \label{eq:3}
\mathrm{A}_{sag}(\mathrm{X}_i) = \sigma(\mathcal{F}_{3\times3}(\mathrm{P}_{max} \circ \mathrm{P}_{avg})), 
\end{equation}

\noindent
where $\sigma$ denotes the sigmoid function, $\mathcal{F}_{3\times3}$ is $3\times3$ \texttt{conv} layer and $\circ$ represents concatenate operation.
Finally, the attention guided feature map $\mathrm{X}_{sag} \in \mathbb{R}^{C\times W \times H}$ is computed as:

\begin{equation} \label{eq:4}
\mathrm{X}_{sag} = \mathrm{A}_{sag}(\mathrm{X}_i) \otimes \mathrm{X}_i,
\end{equation}

\noindent
where $\otimes$ denotes element-wise multiplication. After then, a $2\times2$ \texttt{deconv} upsamples the spatially attended feature map to $28\times28$ resolution. Lastly, a $1\times1$ \texttt{conv} is applied for predicting class-specific masks.

\subsection{VoVNetV2 backbone}
In this section, we propose more effective backbone networks, \textit{VoVNetV2}, for further boosting the performance of CenterMask.
VoVNetV2 is improved from VoVNet~\cite{lee2019energy} by adding residual connection~\cite{he2016deep} and the proposed effective Squeeze-and-Excitation (eSE) attention module to the VoVNet.
VoVNet is a computation and energy efficient backbone network that can efficiently present diversified feature representation because of One-Shot Aggregation (OSA) modules. 
As shown in Figure \ref{fig:osa}(a) OSA module consists of consecutive \texttt{conv} layers and aggregates the subsequent feature maps at once, which can capture diverse receptive fields efficiently and in turn outperforms DenseNet and ResNet in terms of accuracy and speed.

\noindent
\textbf{Residual connection:} Even with its efficient and diverse feature representation, VoVNet has a limitation in terms of optimization.  
As OSA modules are stacked (i.g., deeper) in VoVNet, we observe the accuracy of the deeper models is saturated or degradation.
Specifically, Table ~\ref{tab:vovnet} shows the accuracy of VoVNetV1-99 is lower than that of VoVNetV1-57.
Based on the motivation of ResNet~\cite{he2016deep}, We conjecture that stacking OSA modules make the backpropagation of gradient gradually hard due to the increase of transformation functions such as \texttt{conv}.
Therefore, as shown in Figure~\ref{fig:osa}(b), we also add the identity mapping~\cite{he2016deep} to OSA modules.
Correctly, the input path is connected to the end of an OSA module that is able to backpropagate the gradients of every OSA module in an end-to-end manner on each stage like ResNet.
Boosting the performance of VoVNet, the identity mapping also makes the VoVNet possible to enlarge its depth such as VoVNet-99.

\noindent
\textbf{Effective Squeeze-Excitation (eSE):}
For further boosting the performance of VoVNet, We also propose a channel attention module, effective Squeeze-Excitation (eSE), improving original SE~\cite{hu2018squeeze} more \textit{effectively}.
Squeeze-Excitation (SE)~\cite{hu2018squeeze}, a representative channel attention method adopted in CNN architectures, explicitly models the interdependency between the channels of feature maps to enhance its representation.
The SE module squeezes the spatial dependency by global average pooling to learn a channel specific descriptor and then two fully-connected (FC) layers followed by a sigmoid function are used to rescale the input feature map to highlight only useful channels. In short, given input feature map $\mathrm{X}_i \in \mathbb{R}^{C \times W \times H}$, the channel attention map $\mathrm{A_{ch}}(\mathrm{X_i}) \in \mathbb{R}^{C \times 1 \times 1}$ is computed as:

\begin{equation} \label{eq:5}
\mathrm{A}_{ch}(\mathrm{X}_i) = \sigma(\mathrm{W}_{C}(\delta(\mathrm{W}_{C/r}(\mathcal{F}_{gap}(\mathrm{X}_i)))),
\end{equation}

\noindent
where $\mathcal{F}_{gap}(\mathrm{X}) = \frac{1}{WH}\sum_{i,j=1}^{W,H}\mathrm{X}_{i,j}$ is channel-wise global average pooling, $\mathrm{W}_{C/r}, \mathrm{W}_{C} \in \mathbb{R}^{C \times 1 \times 1}$ are weights of two fully-connected layers, $\delta$ denotes ReLU non-linear operator and $\sigma$ indicates sigmoid function.

However, it is assumed that the SE module has a limitation: channel information loss due to dimension reduction. 
For avoiding high model complexity burden, two FC layers of the SE module need to reduce channel dimension.
Specifically, while the first FC layer reduces input feature channels $C$ to $C/r$ using reduction ratio $r$, the second FC layer expands the reduced channels to original channel size $C$.
As a result, this channel dimension reduction causes channel information loss.

Therefore, we propose effective SE (eSE) that uses only one FC layer with $C$ channels instead of two FCs without channel dimension reduction, which rather maintains channel information and in turn improves performance. the eSE process is defined as:

\begin{equation} \label{eq:6}
\mathrm{A}_{eSE}(\mathrm{X}_{div}) = \sigma(\mathrm{W}_{C}(\mathcal{F}_{gap}(\mathrm{X}_{div}))),
\end{equation}
\begin{equation} \label{eq:7}
\mathrm{X}_{refine} = \mathrm{A}_{eSE}(\mathrm{X}_{div}) \otimes \mathrm{X}_{div},
\end{equation}

\noindent
where $\mathrm{X}_{div} \in \mathbb{R}^{C \times W \times H}$ is the diversified feature map computed by $1\times1$ \texttt{conv} in OSA module.
As a channel attentive feature descriptor, the $\mathrm{A}_{eSE} \in \mathbb{R}^{C \times 1 \times 1}$ is applied to the diversified feature map $\mathrm{X}_{div}$ to make the diversified feature more informative. Finally, when using the residual connection, the input feature map is element-wise added to the refined feature map $\mathrm{X}_{refine}$.
The details of How the eSE module is plugged into the OSA module are shown in Figure \ref{fig:osa}(c).


\subsection{Implementation details}
Since CenterMask is built on FCOS~\cite{Tian_2019_ICCV} object detector, we follow hyper-parameters of the FCOS~\cite{Tian_2019_ICCV} except for positive score threshold 0.03 instead of 0.05 Since FCOS~\cite{Tian_2019_ICCV} does not generate positive RoI samples well in initial training time.
While using FPN levels 3 through 7 with 256 channels in the detection step, we use P3 $\sim$ P7 in the masking step, as mentioned in \ref{sec:3.2}.
We also use mask scoring~\cite{huang2019mask} that recalibrates classification score considering predicted mask quality~(e.g., mask IoU) in Mask R-CNN.

\noindent
\textbf{CenterMask-Lite:}
To achieve real-time processing, we try to make the proposed CenterMask lightweight.
We downsize three parts: backbone, box head, and mask head.
In the backbone, first, we reduce the channels $C$ of FPN from 256 to 128, which can decrease the output of $3\times3$ \texttt{conv} in FPN but also input dimension of box and mask head.
And then, we replace the backbone network with more lightweight VoVNetV2-19 that has 4 OSA modules on each stage comprised of 3 \texttt{conv} layers instead of 5 as in VoVNetv2-39/57.
In the box head, there are four $3\times3$ \texttt{conv} layers with 256 channels on each classification and box branch where the centerness branch is shared with the box branch.
We reduce the number of \texttt{conv} layer from 4 to 2 with 128 channels.
Lastly, in the mask head, we also reduce the number of \texttt{conv} layers and channels in the feature extractor and mask scoring part from (4, 256) to (2, 128), respectively.

\noindent
\textbf{Training:}
We set the number of detection boxes from the FCOS~\cite{Tian_2019_ICCV} to 100, and the highest-scoring boxes are fed into the SAG-mask branch for training mask branch.
We use the same mask target as Mask R-CNN that is made by the intersection between an RoI and its associated ground-truth mask.
During training time, we define a multi-task loss on each RoI as:

\begin{equation} \label{eq:8}
\mathcal{L} = \mathcal{L}_{cls} + \mathcal{L}_{center} + \mathcal{L}_{box} + \mathcal{L}_{mask},
\end{equation}

\noindent
where the classification loss $\mathcal{L}_{cls}$, centerness loss $\mathcal{L}_{center}$, and box regression loss $\mathcal{L}_{box}$ are same as those in ~\cite{Tian_2019_ICCV} and $\mathcal{L}_{mask}$ is the average binary cross-entropy loss identical as in ~\cite{he2017mask}.
Unless specified, the input image is resized to have 800 pixels~\cite{lin2017feature} along the shorter side and their longer side less or equal to 1333.
We train CenterMask by using Stochastic Gradient Descent~(SGD) for 90K iterations~($\sim$12 epoch) with a mini-batch of 16 images and initial learning rate of 0.01 which is decreased by a factor of 10 at 60K and 80K iterations, respectively.
We use a weight decay of 0.0001 and a momentum of 0.9, respectively.
All backbone models are initialized by ImageNet pre-trained weights.

\noindent
\textbf{Inference:}
At test time, the FCOS detection part yields 50 high-score detection boxes, and then the mask branch uses them to predict segmentation masks on each RoI.
CenterMask/CenterMask-Lite use a single scale of 800/600 pixels for the shorter side, respectively.


\begin{figure*}[t]
\centering
   \includegraphics[width=\textwidth]{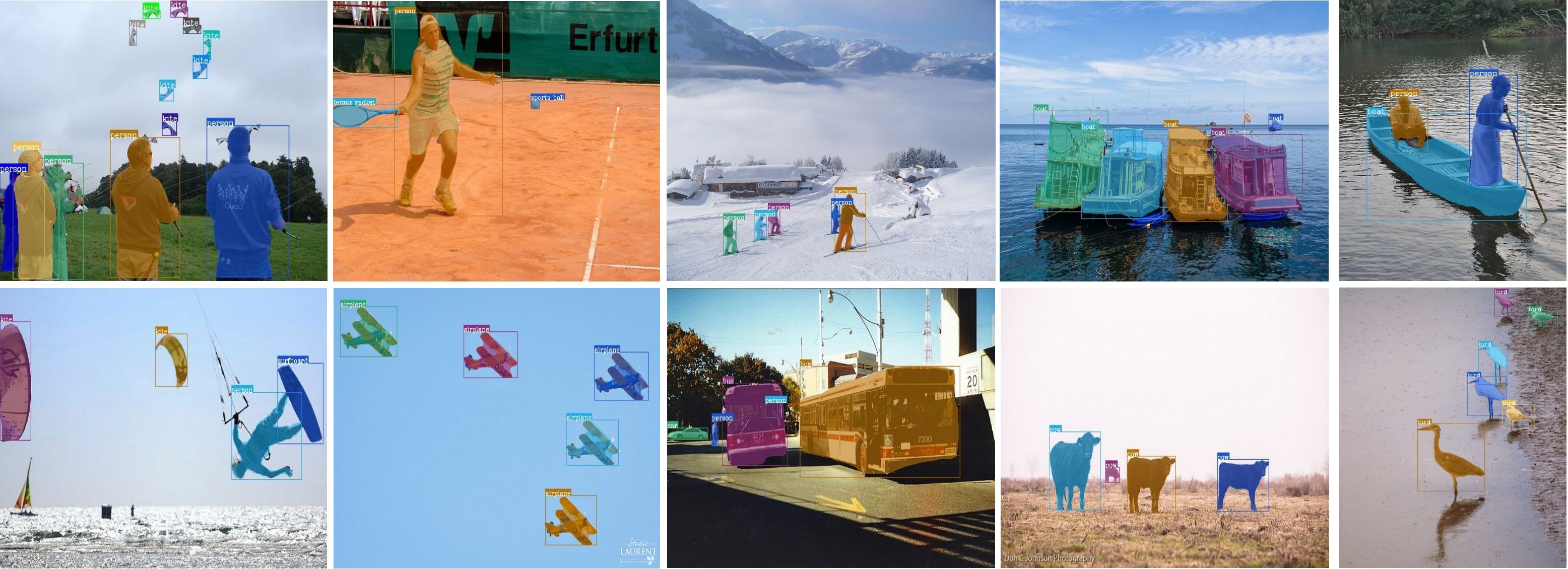} 
\caption{Results of CenterMask with VoVNetV2-99 on COCO \texttt{test-dev2017}.}
\label{fig:vis}
\end{figure*}

\section{Experiments}
In this section, we evaluate the effectiveness of CenterMask on COCO~\cite{lin2014microsoft} benchmarks. All models are trained on the \texttt{train2017} and \texttt{val2017} are used for ablation studies. Final results are reported on \texttt{test-dev} for comparison with state-of-the-arts.
We use AP\textsuperscript{mask} as \textit{mask} average precision AP~(averaged over IoU thresholds), AP\textsubscript{S}, AP\textsubscript{M}, and AP\textsubscript{L} (AP at different scale). We also denote \textit{box} AP as AP\textsuperscript{box}.
All ablation studies are conducted using CenterMask with ResNet-50-FPN.
We report the inference time of models using one \textit{thread} (1 batch size) on the same enviroment equipped with Titan Xp GPU, CUDA v10.0, cuDNN v7.3, and pytorch1.1.
The Qualitative results of CenterMask are shown in Figure \ref{fig:vis}.

\begin{table}[t]
      \centering
      \scalebox{0.9}{
         \begin{tabular}[t]{l|c|c|c}
             Component & AP\textsuperscript{mask} & AP\textsuperscript{box} & Time (ms) \\
             \specialrule{.1em}{.05em}{.05em}
             FCOS~(baseline), \textit{ours} & -     & 37.8  & 57 \\ \hline
             + mask head (Eq.~\ref{eq:1}~\cite{lin2017feature}) & 33.4  & 38.3  & 67 \\
             + mask head (Eq.~\ref{eq:2}, \textit{ours}) & 33.8  & 38.7  & 67 \\
             + SAM & 34.0  & 38.9  & 67 \\
             + Mask scoring & 34.7  & 38.8  & 72 \\
          \end{tabular}%
      }
        \caption{\textbf{Spatial Attention Guided Mask~(SAG-Mask)}\\These models use ResNet-50 backbone. We note that the mask heads with Eq.\ref{eq:1} is same as the mask branch of Mask R-CNN. SAM and Scoring denotes the proposed Spatial Attention Module and mask scoring~\cite{
        huang2019mask}.  \mbox{} }
        \label{tab:mask}%
        \vspace{-0.3cm}
\end{table}

\subsection{Ablation study}
\noindent
\textbf{Scale-adaptive RoI assignment function:}
Comparing to Equation \ref{eq:1}~\cite{lin2017feature}, we validate the proposed Equation \ref{eq:2} in CenterMask.
Table \ref{tab:mask} shows that our scale-adaptive RoI assignment function considering the input scale improves by 0.4\% AP\textsuperscript{mask} over the counterpart.
It means that Equation \ref{eq:2} regarding the ratio of input/RoI is more scale-adaptive than Equation \ref{eq:1}.

\noindent
\textbf{Spatial Attention Guided Mask:}
Table~\ref{tab:mask} demonstrates the influence of each component in building Spatial Attention Guided Mask~(SAG-Mask).
The baseline, FCOS~\cite{Tian_2019_ICCV} object detector, starts from 37.8\% AP\textsuperscript{box} with the run time of 57 ms.
Adding only naive mask head improves the box performance by 0.5\% AP\textsuperscript{box} and obtains 33.4\% AP\textsuperscript{mask}.
With the prementioned scale-adaptive RoI mapping strategy, our spatial attention module, \textit{SAM}, makes the mask performance forward because the spatial attention module helps the mask predictor to focus on informative pixels but also suppress noise.
It can also be seen that the detection performance is boosted when using SAM.
We suggest that result from the SAM, the refined feature maps of mask head would also have a secondary effect on the detection branch that shares feature maps of the backbone.

The SAG-mask also deploys the mask scoring ~\cite{huang2019mask} that recalibrates the score regarding the predicted mask IoU.
As a result, the mask scoring increases performance by 0.7\% AP\textsuperscript{mask}.
We note that the mask scoring cannot boost detection performance because the recalibrated mask score adjusts the ranks of mask results in the evaluation step, not refines the features of the mask head like the SAM.
Besides, SAM rarely causes extra computation while the mask scoring leads to computation overhead~(e.g., +5ms).
\\

\begin{table}[t]
     \centering
     \scalebox{0.95}{
       \begin{tabular}{c|c|c}
       Feature Level & AP\textsuperscript{mask} & AP\textsuperscript{box} \\
       \specialrule{.1em}{.05em}{.05em}   
       P3 $\sim$ P7 & 34.4  & 38.8 \\
       P3 $\sim$ P6 & 34.6  & 38.8 \\
       P3 $\sim$ P5 & \textbf{34.6}  & \textbf{38.9} \\
       P3 $\sim$ P4 & 34.4  & 38.5 \\
       \end{tabular}%
      }
     \caption{Feature level ranges for RoIAlign~\cite{he2017mask} in CenterMmask. P3$\sim$P7 denotes the feature maps with output stride of $2^3 \sim 2^7$  }
     \label{tab:level}%
     \vspace{-0.3cm}
\end{table}

\begin{table*}[t]
\centering
\scalebox{1.0}{
\begin{tabular}{l|c|cccc|cccc|c}
Backbone    & Params. & AP\textsuperscript{mask}   & AP$^\text{mask}_{S}$  & AP$^\text{mask}_{M}$  & AP$^\text{mask}_{L}$  & AP\textsuperscript{box} & AP$^\text{box}_{S}$ & AP$^\text{box}_{M}$ & AP$^\text{box}_{L}$ & Time (ms) \\ \specialrule{.1em}{.05em}{.05em}
MobileNetV2~\cite{sandler2018mobilenetv2} & 28.7M   & 29.5 & 12.0   & 31.4 & 43.8 & 32.6                 & 17.8                  & 35.2                  & 43.2                  & \textbf{56}        \\
VoVNetV2-19~\cite{lee2019energy} & 37.6M   & \textbf{32.2} &\textbf{14.1}      &\textbf{34.8}      &\textbf{48.1}      & \textbf{35.9}                 &\textbf{20.8}                       &\textbf{39.2}                       &\textbf{47.6}                       & 59      \\ \hline
HRNetV2-W18~\cite{sun2019high} & 36.4M   &33.0   & 14.3 & 34.7 & 49.9 & 36.7                 & 20.7                  & 39.4                  & 49.3                  & 80        \\
ResNet-50~\cite{he2016deep}   & 51.2M   & 34.7 &15.5      &37.6      &51.5      & 38.8                 &22.4                       &42.5                       &51.1                       & 72        \\
VoVNetV1-39~\cite{lee2019energy} & 49.0M   & 35.3   &15.5      &38.4      &52.1      & 39.7                 &23.0                       &43.3                       &52.7                       & \textbf{68}        \\
VoVNetV2-39 & 52.6M   & \textbf{35.6} &\textbf{16.0}      &\textbf{38.6}      &\textbf{52.8}      & \textbf{40.0}                   &\textbf{23.4}                       &\textbf{43.7}                       &\textbf{53.9}                       & 70        \\ \hline
HRNetV2-W32~\cite{sun2019high} & 56.2M   & 36.2 & 16.0   & 38.4 & 53.0   & 40.6                 & 23.0                    & 43.8                  & 53.1                  & 95        \\
ResNet-101\cite{he2016deep}  & 70.1M   & 36.0   &16.5      &39.2      &54.4      & 40.7                 &23.4                       &44.3                       &54.7                       & 91        \\
VoVNetV1-57~\cite{lee2019energy} & 63.0M   &36.1      &16.2      &39.2      &54.0      &40.8                      &23.7                       &44.2                       &55.3                       &\textbf{74}           \\
VoVNetV2-57 & 68.9M   & \textbf{36.6} &\textbf{16.9}      &\textbf{39.8}      &\textbf{54.5}      & \textbf{41.5}                 &\textbf{24.1}                       &\textbf{45.2}                       &\textbf{55.2}                       & 76        \\ \hline
HRNetV2-W48~\cite{sun2019high} & 92.3M   & 38.1 & 17.6   & 41.1 & 55.7   & 43.0                 & 25.8                    & 46.7                  & 55.9                  & 126        \\
ResNeXt-101~\cite{xie2017aggregated} & 114.3M  & 38.3 & \textbf{18.4} & 41.6 & 55.4 & 43.1                 & \textbf{26.1}                  & 46.8                  & 55.7                  & 157       \\
VoVNetV1-99~\cite{lee2019energy} &  83.6M  & 31.5 & 13.5   & 33.5 & 46.5   & 35.3                 & 19.7                  & 38.1                  & 46.6                  & \textbf{101} \\       
VoVNetV2-99 & 96.9M   & \textbf{38.3} & 18.0   & \textbf{41.8} & \textbf{56.0}   & \textbf{43.5}                 & 25.8                  & \textbf{47.8}                  & \textbf{57.3}                  & 106      
\end{tabular}
}
  \caption{\textbf{CenterMask with other backbones} on COCO \texttt{val2017}. Note that all mdoels are trained with a same manner~(e.g., 12 epoch, 16 batch size, without train \& test augmentation). The inference time is reported on same Titan Xp GPU.}
  \label{tab:backbone}%
   \vspace{-0.4cm}
\end{table*}

\begin{table}[t]
     \centering
     \scalebox{0.85}{
       \begin{tabular}{l|c|c|c|c}
       Backbone & Params. & AP\textsuperscript{mask}    & AP\textsuperscript{box}  & Time (ms) \\
       \specialrule{.1em}{.05em}{.05em}
       VoVNetV1-39 & 49.0M & 35.3  & 39.7  & 68 \\
       + residual & 49.0M & 35.5~\color{blue}{(+0.2)}  & 39.8~\color{blue}{(+0.1)}  & 68 \\
       + SE~\cite{hu2018squeeze}   & 50.8M & 34.6~\color{red}{(-0.7)}  & 39.0~\color{red}{(-0.7)}  & 70 \\
       + eSE,~\textit{ours}  &  52.6M  & 35.6~\color{blue}{(+0.3)}  & 40.0~\color{blue}{(+0.3)}  & 70 \\
       \hline
       VoVNetV1-57 &  63.0M  & 36.1  & 40.8  & 74 \\
       + residual &  63.0M  & 36.4~\color{blue}{(+0.3)}  & 41.1~\color{blue}{(+0.3)}  & 74 \\
       + SE~\cite{hu2018squeeze}   &  65.9M  & 35.9~\color{red}{(-0.2)}  & 40.8  & 77 \\
       + eSE,~\textit{ours}  &  68.9M  & 36.6~\color{blue}{(+0.5)}  & 41.5~\color{blue}{(+0.7)}  & 76 \\
       \hline
       VoVNetV1-99 &  83.6M  & 31.5  & 35.3  & 101 \\
       + residual &  83.6M  & 37.6~\color{blue}{(+6.1)}  & 42.5~\color{blue}{(+7.2)}  & 101 \\
       + SE~\cite{hu2018squeeze}   &   88.0M & 37.1~\color{blue}{(+5.6)}  & 41.9~\color{blue}{(+6.6)}  & 107 \\
       + eSE,~\textit{ours}  &  96.9M  & 38.3~\color{blue}{(+6.8)}  & 43.5~\color{blue}{(+8.2)}  & 106 \\
       \end{tabular}%
      }
     \caption{\textbf{VoVNetV2} Start from VoVNetV1, VoVNetV2 is improved by adding residual connection~\cite{he2016deep} and the proposed effetive SE (eSE). }
     \label{tab:vovnet}%
     \vspace{-0.3cm}
\end{table}

\noindent
\textbf{Feature selection.}
We also ablate which feature level range is suitable for our CenterMask based one-stage detector.
Since FCOS~\cite{Tian_2019_ICCV} detector extract features from P3 $\sim$ P7, we start the same feature levels in the SAG-mask branch.
As shown in Table \ref{tab:level}, the performance of the P3 $\sim$ P7 range is not as good as other ranges.
We speculate P7 feature map is too small to extract fine features for pixel-level prediction~(e.g., $7\times7$).
We observe that P3 $\sim$ P5 feature range achieves the best result, which means feature maps with a bigger resolution are advantageous for the mask prediction.
\\

\noindent
\textbf{VoVNetV2:}
We extend VoVNet to VoVNetV2 by using residual connection and the proposed effective SE~(eSE) module into the VoVNet.
Table~\ref{tab:vovnet} shows residual connection consistently improves VoVNet-39/57/99.
In particular, the reason that the improved AP margin of VoVNet-99 is bigger than VoVNet-39/57 is that VoVNet-99 comprised of more OSA modules can have more effect of residual connection that alleviates the optimization problem.

To validate eSE, we also apply the SE~\cite{hu2018squeeze} to the VoVNet and compare it with the proposed eSE.
As shown in Table ~\ref{tab:vovnet}, the SE worsens the performance of VoVNet or has no effect because the diversified feature map of OSA module losses channel information due to channel dimension reduction in the SE.
Contrary to the SE, our eSE maintaining channel information using only 1 FC layer boosts both AP\textsubscript{mask} and AP\textsubscript{box} from VoVNetV1 with slight computation.
\\

\begin{table*}[t]
  \centering
  \scalebox{0.85}{
    \begin{tabular}{l|l|c|cccc|cccc|c|c|c}
    \multicolumn{1}{c|}{Method} & \multicolumn{1}{c|}{Backbone} & epochs & AP\textsuperscript{mask}   &AP$^\text{mask}_{S}$  & AP$^\text{mask}_{M}$  & AP$^\text{mask}_{L}$  & AP\textsuperscript{box} & AP$^\text{box}_{S}$ & AP$^\text{box}_{M}$ & AP$^\text{box}_{L}$ & Time & FPS &GPU \\
    \specialrule{.1em}{.05em}{.05em}
    Mask R-CNN, \textit{ours} & R-101-FPN & 24    &37.9       &\textbf{18.1}       &40.3       &53.3       &42.2       &24.9       &45.2       &52.7       &94   &10.6       & V100 \\
    ShapeMask~\cite{Kuo_2019_ICCV} & R-101-FPN & N/A   &       37.4  &       16.1  &       40.1  &       53.8  &       42.0  &       24.3  &       45.2  &       53.1  &        125 & 8.0  & V100 \\
    TensorMask~\cite{Chen_2019_ICCV} & R-101-FPN & 72    &       37.1  &       17.4  &       39.1  &       51.6  &-       &   -    &   -    &    -   &        380 & 2.6  & V100 \\
    RetinaMask~\cite{fu2019retinamask} & R-101-FPN & 24    &       34.7  &       14.3  &       36.7  &       50.5  &       41.4  &       23.0  &       44.5  &       53.0  &98    & 10.2        & V100 \\
    \textbf{CenterMask} & R-101-FPN & 24    &       \textbf{38.3}  &       17.7  &       \textbf{40.8}  &       \textbf{54.5}  &       \textbf{43.1}  &       \textbf{25.2}  &       \textbf{46.1}  &       \textbf{54.4}  &         \textbf{72}    &\textbf{13.9}   & V100 \\
    \hline
    \textbf{CenterMask*} & R-101-FPN & 36    &       39.8  &       21.7  &       42.5  &       52.0  &       44.0  &       25.8  &       46.8  &       54.9  &         66    &15.2   & V100 \\
    \hline
    Mask R-CNN, \textit{ours} & X-101-FPN & 36    &39.3       &19.8       &41.4       &55.0       &44.1       &27.0       &46.7       &54.6       & 165  &   6.1    & V100 \\
    \textbf{CenterMask} & X-101-FPN & 36    &   39.6   &   19.7    &   42.0    &   55.2    &    44.6   &   27.1    &   47.2    &  55.2     &        123  &8.1 & V100 \\
    \textbf{CenterMask} & V-99-FPN & 36    &    \textbf{40.6}   &   \textbf{20.1}    &   \textbf{42.8}    &   \textbf{57.0}    &   \textbf{45.8}    &   \textbf{27.8}    &   \textbf{48.3}    &   \textbf{57.6}    &         \textbf{84}  &\textbf{11.9} & V100 \\
    \hline
    \textbf{CenterMask*} & V-99-FPN & 36    &    41.8   &   24.4    &   44.4    &   54.3    &   46.5    &   28.7    &   48.9    &   57.2    &         77  &12.9 & V100 \\
    \specialrule{.1em}{.05em}{.05em}
    YOLACT-400~\cite{Bolya_2019_ICCV} & R-101-FPN & 48    &       24.9  &         5.0  &       25.3  &       \textbf{45.0}  &       28.4  &       10.7  &       28.9  &       \textbf{43.1}  &           22 & 45.5  & Xp \\
    \textbf{CenterMask-Lite} & M-v2-FPN & 48    & \textbf{26.7} &         \textbf{9.0}  &       \textbf{27.0}  & 40.9  & \textbf{30.2} & \textbf{14.2}    & \textbf{31.9}  & 40.9  & \textbf{20} & \textbf{50.0}  & Xp \\
    \hline
    YOLACT-550~\cite{Bolya_2019_ICCV} & R-50-FPN & 48    &       28.2  &         9.2  &       29.3  &       44.8  &       30.3  &       14.0  &       31.2  &       43.0  &         23&43.5   & Xp \\
    \textbf{CenterMask-Lite} & V-19-FPN & 48    & \textbf{32.4} &       \textbf{13.6}  &       \textbf{33.8}  & \textbf{47.2}  & \textbf{35.9} & \textbf{19.6}    & \textbf{38.0}  & \textbf{45.9}  & \textbf{23}&\textbf{43.5} & Xp \\
    \hline
    YOLACT-550~\cite{Bolya_2019_ICCV} & R-101-FPN & 48    &       29.8  &         9.9  &       31.3  &       47.7  &       31.0  &       14.4  &       31.8  &       43.7  &         30 &33.3  & Xp \\
    YOLACT-700~\cite{Bolya_2019_ICCV} & R-101-FPN & 48    &       31.2  &       12.1  &       33.3  &       47.1  &       33.7  &       16.8  &       35.6  &       45.7  &         42&23.8  & Xp \\
    \textbf{CenterMask-Lite} & R-50-FPN & 48    &32.9       &12.9       &34.7       &48.7       &36.7       &18.7       &39.4       &48.2       &29   &34.5       & Xp \\
    \textbf{CenterMask-Lite} & V-39-FPN & 48    &\textbf{36.3}       &\textbf{15.6}       &\textbf{38.1}       &\textbf{53.1}       &\textbf{40.7}       &\textbf{22.4}       &\textbf{43.2}       &\textbf{53.5}       &\textbf{28}   & \textbf{35.7}       & Xp \\
    \end{tabular}%
   }
  \vspace{-0.3cm}
  \caption[Caption for LOF]{\textbf{CenterMask} instance segmentation and detection performance on COCO \texttt{test-dev2017}. Mask R-CNN, RetinaMask, and CenterMask are implemented on the same base code~\cite{massa2018mrcnn} and CenterMask* is implemented on top of \texttt{Detectron2}\protect\footnotemark~\cite{wu2019detectron2}. R, X, V, and M denote ResNet, ResNeXt-32x8d, VoVNetV2, and MobileNetV2, respectively. For fair compariosn, these results are tested with \textit{one thread} and single-scale.}
  \label{tab:SOTA}
  \vspace{-0.35cm}
\end{table*}%

\noindent
\textbf{Comparison to other backbones:}
We expand VoVNetV2 on various scales; large~(V-99), base~(V-39/57), and lightweight~(V-19) which correspond to ResNeXt-32-8d~\cite{xie2017aggregated} \& HRNet-W48~\cite{sun2019high}, ResNet-50/101~\cite{he2016deep} \& HRNet-W18/W32~\cite{sun2019high}, and MobileNetV2~\cite{sandler2018mobilenetv2}, respectively.
Table~\ref{tab:backbone} and Figure \ref{fig:FPS}~(bottom) demonstrate VoVNetV2 is well-balanced backbone network in terms of accuracy and speed.
While VoVNetV1-39 already outperforms its counterparts, VoVNetV2-39 shows better performance than ResNet-50/HRNet-W18 by a large margin of 1.2\%/2.6\% at faster speeds, respectively.
Especially, the gain of AP\textsuperscript{box} is bigger than AP\textsuperscript{mask}, 1.5\%/3.3\%, respectively.
A similar result pattern is shown in VoVNetV2-57 with its counterparts.

For large model, showing much faster run time~($1.5\times$), VoVNetV2-99 achieves competitive AP\textsuperscript{mask} or higher AP\textsuperscript{box} than ResNeXt-101-32x8d despite fewer model parameters.
For small model, VoVNetV2-19 outperforms MobileNetV2 by a large margin of 1.7\% AP\textsuperscript{mask}/3.3\%AP\textsuperscript{box}, with comparable speed.

\subsection{Comparison with state-of-the-arts methods}\label{sec:SOTA}
For further validation of the CenterMask, we compare the proposed CenterMask with state-of-the-art instance segmentation methods.
As most methods~\cite{liu2016ssd,Chen_2019_ICCV,He_2019_ICCV,Bolya_2019_ICCV,fu2019retinamask} use train augmentation, we also adopt the scale-jitter where the shorter image side is randomly sampled from [640, 800] pixels~\cite{He_2019_ICCV}. For Centermask-Lite, [580, 600] scale jittering is used for training.
We train CenterMask and CenterMask-Lite for 24/36 epochs and 48 epochs, respectively.
Note that we do not use test-time augmentation~\cite{He_2019_ICCV}~(multi-scale).
The other hyper-parameters are kept same as ablation study.
For fair speed comparison, we inference models on the same GPU as counterparts.
Specifically, since most large models are tested on V100 GPU and YOLACT~\cite{Bolya_2019_ICCV} models are reported on Titan Xp GPU, we also report CenterMask models on V100 and CenterMask-Lite models on Xp.

Under the same ResNet-101 backbone, CenterMask outperforms all other counterparts in terms of both accuracy~(AP\textsuperscript{mask}, AP\textsuperscript{box}) and speed.
In particular, compared to RetinaMask~\cite{fu2019retinamask} that has similar architecture~(i.g., one-stage detector + mask branch), CenterMask achieves 3.6\%AP\textsuperscript{mask} gain.
In less than half training epochs, CenterMask also surpasses the dense sliding window method, TensorMask~\cite{Chen_2019_ICCV}, by 1.2\%AP\textsuperscript{mask} at $\times5$ faster speed.
Furthermore, to the best of our knowledge, the CenterMask with VoVNetV2-99 is the first method to achieves 40\% AP\textsuperscript{mask} at over 10 fps.
It is noted that after first submission, \texttt{Detectron2}~\cite{wu2019detectron2} has been released that is a better baseline code.
Thus, we also re-implement our CenterMask* on top of \texttt{Detectron2}~\cite{wu2019detectron2} and obtain further performance gain.

We also compare with YOLACT~\cite{Bolya_2019_ICCV} that is the representative real-time instance segmentation.
We use four kinds of backbones (e.g., MobileNetV2, VoVNetV2-19, VoVNetV2-39, and ResNet-50), which have a different accuracy-speed tradeoff.
Table \ref{tab:SOTA} and Figure \ref{fig:FPS} (top) demonstrate CenterMask-Lite is consistently superior to YOLACT in terms of accuracy and speed.
Compared to YOLACT, all CenterMask-Lite models achieve over 30 fps speed with large margins of both AP\textsuperscript{mask} and AP\textsuperscript{box}.

\footnotetext{After the initial submission, \texttt{Detectron2}~\cite{wu2019detectron2} has been released and we has developed the improved CenterMask* on top of the \texttt{Detectron2}~\cite{wu2019detectron2}.}
\section{Discussion}
\noindent
In Table ~\ref{tab:SOTA}, we observe that using the same ResNet-101 backbone, Mask R-CNN~\cite{he2017mask} shows better performance than CenterMask on small object.
We conjecture that Mask R-CNN~\cite{he2017mask} uses larger feature maps~(P2) than CenterMask~(P3) in which the mask branch can extract much finer spatial layout of an object than the P3 feature map.
We note that there are still rooms for improving one-stage instance segmentation performance like techniques~\cite{cai2018cascade,chen2019hybrid} of Mask R-CNN~\cite{he2017mask}.


\section{Conclusion}
We have proposed a real-time anchor-free one-stage instance segmentation and more effective backbone networks.
Adding spatial attention guided mask branch to the anchor-free one stage instance detection, CenterMask achieves state-of-the-art performance at real-time speed.
The newly proposed VoVNetV2 backbone spanning from lightweight to larger models makes CenterMask well-balanced performance in terms of speed and accuracy. 
We hope CenterMask will serve as a baseline for real-time instance segmentation. 
We also believe our proposed VoVNetV2 can be used as a strong and efficient backbone network for various vision tasks~\cite{yun2019vision,jo2019sc}.

\noindent
\textbf{Acknowledgements.}
We thank Hyung-il Kim and the DeepView team in ETRI.
This work was supported by Institute of Information \& Communications Technology Planning \& Evaluation (IITP) grant funded by
the Korea government (MSIT) (No. B0101-15-0266, Development of High Performance Visual BigData Discovery Platform for
Large-Scale Realtime Data Analysis and No. 2020-0-00004, Development of Previsional Intelligence based on Long-term Visual Memory
Network).

{\small
\bibliographystyle{ieee_fullname}
\bibliography{ref}
}

\end{document}